\documentclass[final,1p]{elsarticle}

\usepackage{amssymb}
\usepackage{times}
\usepackage{epsfig}
\usepackage{graphicx}
\usepackage{amsmath}
\usepackage{graphicx,subfigure}
\usepackage{graphicx,subfigure}
\usepackage{algorithm}
\usepackage{algorithmic}
\usepackage{multirow}
\usepackage{xcolor}
\usepackage{ifpdf}
\usepackage{epstopdf}

\usepackage[nodots]{numcompress}

\ifpdf
\DeclareGraphicsExtensions{.pdf,.png,.jpg,.mps}
\else
\DeclareGraphicsExtensions{.eps}
\fi

\begin{document}

\begin{frontmatter}



\title{DCI: Discriminative and Contrast Invertible Descriptor}

\author[label2]{\corref{cor1}Zhenwei Miao}
\ead{miaozhenwei@gmail.com}
\author[label2]{Kim-Hui Yap}
\ead{ekhyap@ntu.edu.sg}
\author[label2]{Xudong Jiang}
\ead{exdjiang@ntu.edu.sg}
\author[label2]{Subbhuraam Sinduja}
\ead{sinduja001@ntu.edu.sg}
\author[label3]{Zhenhua Wang}
\ead{zhwang.me@gmail.com}
\address[label2]{School of Electrical and Electronic Engineering,\\
Nanyang Technological University, Singapore 639798}
\address[label3]{Rachel and Selim Benin School of Computer Science and Engineering,\\ 
Hebrew University of Jerusalem, Israel 91905}

\cortext[cor1]{Corresponding author. Tel. +65 9755 0569}


\begin{abstract}
Local feature descriptors have been widely used in fine-grained visual object search thanks to their robustness in scale and rotation variation and cluttered background. However, the performance of such descriptors drops under severe illumination changes. In this paper, we proposed a Discriminative and Contrast Invertible (DCI) local feature descriptor. In order to increase the discriminative ability of the descriptor under illumination changes, a Laplace gradient based histogram is proposed. A robust contrast flipping estimate is proposed based on the divergence of a local region. Experiments on fine-grained object recognition and retrieval applications demonstrate the superior performance of DCI descriptor to others.
\end{abstract}


\end{frontmatter}


\section{Introduction}

\begin{figure}[t]
  \centering
  \includegraphics[width=0.5\textwidth]{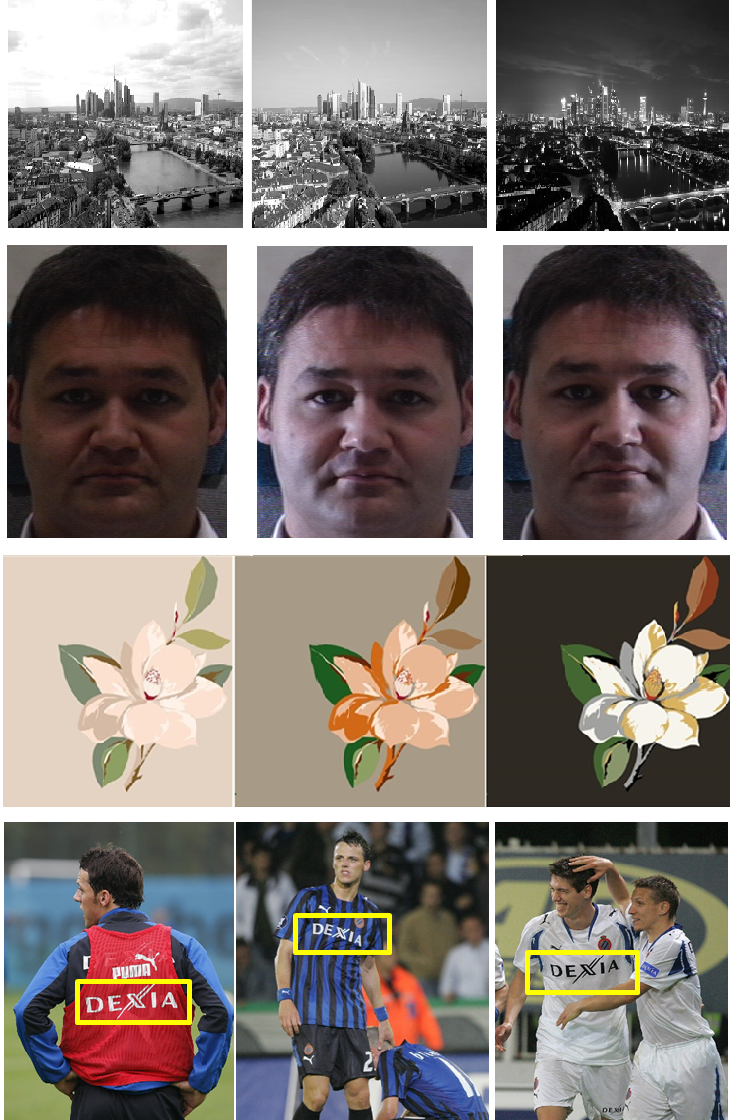}
  \caption{Sample images of commercial designs with different illuminations. The first row shows the same landmarks captured under different whether and lighting, the second rows shows the face with different illumination, the third row shows the wallpaper with different colors and fourth row shows the logos with contrast variation.}
  \label{products_sample_images}
\end{figure}

In the past decades, local feature description \cite{Lowe04,xie2015ride,rublee2011orb,Wang11,abdel2006csift,Zhao13,Tuytelaars08,Mikolajczyk05,Kim20133268,Qi2016420,miao2013median} has drawn intensive attention in various applications, such as image retrieval \cite{Arandjelovic12,verdie2015tilde,zhang2015feature}, image matching  \cite{hauagge2012image}, object search \cite{jiang2015randomized} and face recognition \cite{Geng13}. Although the deep learning based CNN approaches \cite{jia2014caffe,zagoruyko2015learning,simo2015discriminative} are emerging in recent years and   have achieved great success in quite a few domains \cite{girshick2014rich,Arandjelovic12,zeiler2014visualizing,krizhevsky2012imagenet,razavian2014cnn,miao2017laplace}, the advantages of traditional local features, especially in fine-grained object search tasks, are evident given their low computational requirement and robustness to scale, rotation, illumination changes and cluttered background. In general, local patches extracted by various interest point detectors \cite{calonder2010brief,Miao15TIP,Lowe04,Kimmel11,Wang13,Rosten10,Miao13PR,Miao12icassp, Miao13icassp} are expected to be described by descriptors that are discriminative and tolerant to  geometrical and illuminative variations. Despite efforts have been devoted to developing kinds of descriptors, the performances of existing descriptors are still limited especially under conditions where illumination changes greatly. Such changes occur frequently in real-world scenarios. For example, as shown in Fig. \ref{products_sample_images}, the first row shows the same landmark taken under different weather and lighting, the second row shows the face images under different illumination condition, the third row displays the same logo in T-shirts with opposite contrast while the fourth row shows the identical art design pattern implemented in completely different level of illumination. In such cases, previous descriptors encounter difficulties in recognition and retrieving the images with the identical objects yet different illumination conditions.  

There are two types of contrast changes: bright-dark order preserved changes and the bright-dark order disturbed changes. In images with bright-dark order preserved changed, the relative order of the pixel intensity remains the same while the image patch becomes either brighter or darker following linear and/or nonlinear transformations. In contrast, images with bright-dark disturbed changes will not retain the relative order. The majority of previous local feature descriptors, such as SIFT \cite{Lowe04}, GLOH \cite{Mikolajczyk05} and PCA-SIFT \cite{ke2004pca}, are designed for the first type. SIFT is suggested to be the most stable descriptor for common image deformations \cite{Mikolajczyk05}. It consists of a $4\times4$ array of histograms, and each histogram contains eight orientation bins that are weighted by the gradient magnitudes. Because of the location and orientation quantizations, the SIFT descriptor is robust to small geometric distortion and location errors. As an extension of SIFT, the GLOH \cite{Mikolajczyk05} enhances the robustness and discriminative ability by increasing the number of spatial bins and orientation to 16 and 17, respectively. The Principal Component Analysis (PCA) is carried out to reduce the  number of feature dimensions of GLOH to be equivalent to SIFT's. PCA-SIFT \cite{ke2004pca} further reduces the dimensions of SIFT from 128 to 20 by applying PCA to gradient image patches. Although these descriptors are robust to certain variations and distortions, they fail to address the issues caused by severe nonlinear illumination changes  and bright-dark order disturbed changes as abovementioned \cite{Wang11,fan2012rotationally}. 

\begin{figure}[t]
  \centering
  \includegraphics[width=0.8\textwidth]{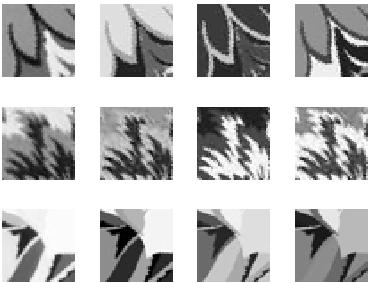}
  \caption{Image patches with different illuminations. Each row shows image patches from the same design.}
  \label{image_patch}
\end{figure}

Recently, alternative local feature descriptors have been proposed to solve the problems caused by illumination changes. Instead of using the gradient orientation and magnitude, the intensity order is adopted by the Local Intensity Order Pattern (LIOP) descriptor \cite{Wang11}. The local image region is divided into several sub-regions according to the intensity order of pixels in that region. Local intensity order patterns are computed to describe each sub-region. As a result, LIOP can identify features with consistent illumination changes in which the order of pixel intensity is preserved. However, it cannot effectively tackle the bright-dark order of the pixel changes that may be caused by noises, illumination direction changes, illumination sources changes, or artificial designs as shown in Fig. \ref{image_patch}. 

The bright-dark disturbed changes are even more complicated than the order preserved ones. To the best of our knowledge, few works \cite{ma2010mi,xie2014max} have been done to solve the bright-dark disturbed changes where both the amplitude and the relative order of the bright/dark pixels may vary. When the gradient orientation changes to its opposite direction with a 180 degree rotation (i.e., bright-dark inversion), the rotated descriptor cannot be matched with the original one. However, the original images and the bright/dark inverted images can be converted by a fixed transfer function \cite{ma2010mi,xie2014max}, allowing the distribution of gradient orientation and amplitudes in a local region to be transfered in a different order. Hence, the mirror and inversion invariant SIFT (MI-SIFT) \cite{ma2010mi} is developed with a canonical form to identify the relationship between descriptors in the original image patches and the illumination/mirror inversed ones. The MI-SIFT is invariant to both the bright and geometric flipping through replacing the corresponding bins in the SIFT descriptor with symmetric bins. The MAX-SIFT \cite{xie2014max} is designed with a different approach from using advanced arithmetic operations; it rearranges the SIFT bins under different flipped versions with the original bins. Both of MI-SIFT and MAX-SIFT descriptors can work for instances of bright and/or geometric flipping. However, they still cannot solve the problem of partial illumination changes as shown in Fig. \ref{image_patch}. Although the Shape Context (SC) \cite{belongie2002shape} using edges instead of image intensities provides a potential ways to describe local regions with partial illumination changes, stable edge detection is challenging and unclear.  
 
In this work, we propose a novel descriptor named Discriminative and Contrast Invertible (DCI) descriptor.  The DCI is designed to enhance the performance of local feature descriptors for large illumination changes and contrast inversions. Unlike the local gradients that are easily affected by illumination conditions, a Laplace gradient we proposed in this paper is effective in capturing local characteristics with illumination variations. The DCI descriptor is formed by concatenating the histograms of the Laplace gradient in different bins.  Moreover, to address the bright/dark inversion problem, we proposed an inversion-estimation function based on the divergence of gradient. The proposed descriptor is evaluated across diverse object visual search tasks, including searching logos, commercial design images and faces. Experimental results demonstrate that the proposed DCI descriptor outperforms the state-of-the-art descriptors.

\section{The Proposed Descriptor}
	
\begin{figure*}[t]
	\centering
	\includegraphics[width=1\textwidth]{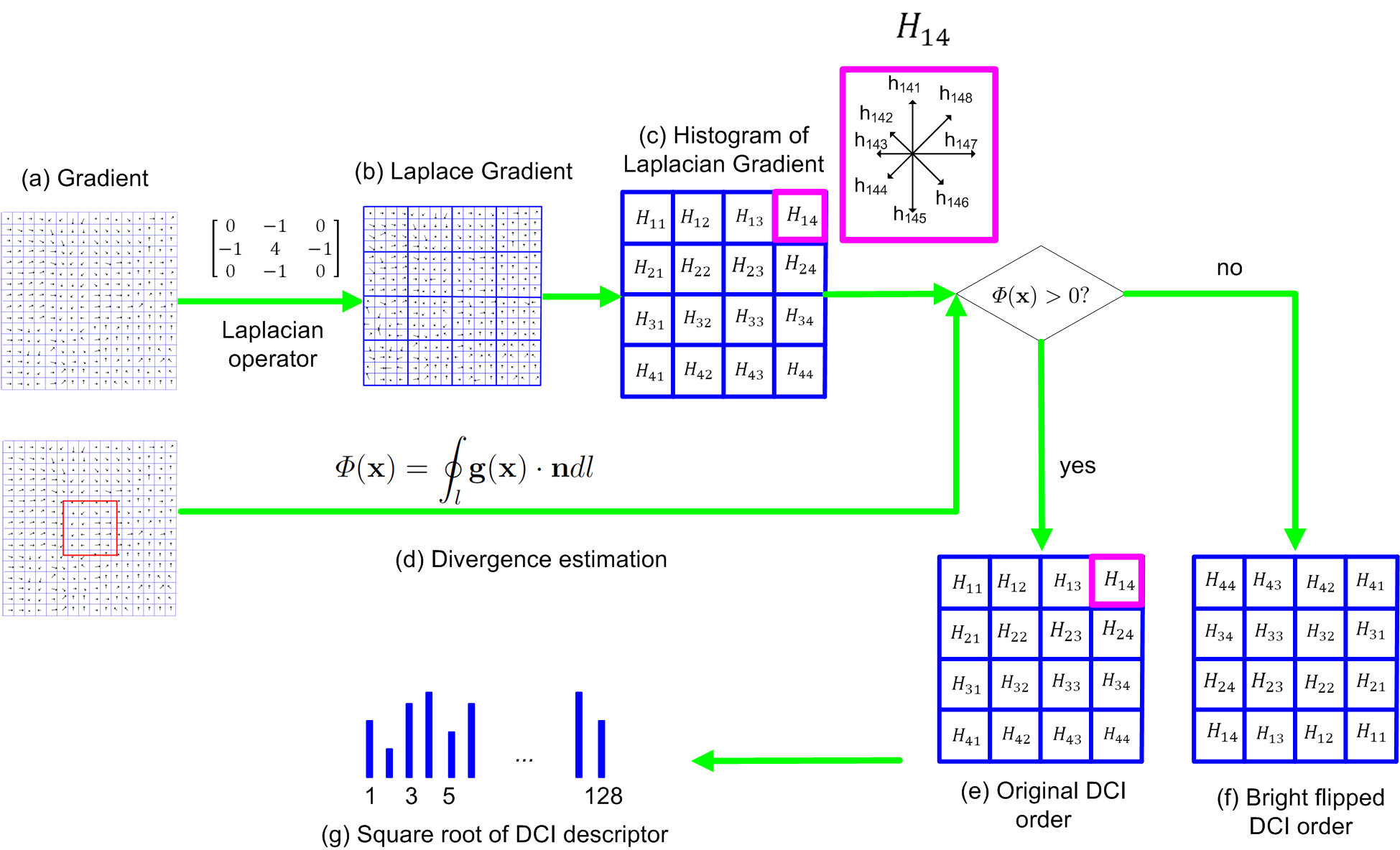}
	\caption{Flowchart of the proposed DCI descriptor. (a)  gradient of image patch, (b)  Laplace gradient, (c) histogram of Laplace gradient in the $4\times4$ blocks, (d) divergence-based brightness flipping estimator, (e) original DCI order, (f) DCI order under brigh-dark flip case, and (g) DCI descriptor after $L_1$ normalization and square rooting.}  \label{flowchart_descriptor}
\end{figure*}
	
The Laplace gradient and the divergence based contrast flipping estimation function are proposed in the DCI descriptor to solve the problems caused by the brightness intensity and bright-dark order changes. Fig. \ref{flowchart_descriptor} illustrates the work flow of the DCI descriptor. As how the SIFT descriptor is developed, a local patch with the standard size of $31\times31$\footnote{Here we follow the default of the SIFT descriptor provided in vlfeat http://www.vlfeat.org/} is extracted around each interest point at the given scale and aligned along its dominant orientation \cite{Lowe04}. Following that, the gradient of the image patch is extracted as shown in Fig. \ref{flowchart_descriptor}(a). The Laplace gradient shown in Fig. \ref{flowchart_descriptor}(b) is computed to enhance the discriminative ability and robustness to brightness changes. Then, the Laplace gradient map in Fig. \ref{flowchart_descriptor}(b) is divided into $4\times4$ sub-regions. The distribution of the Laplacian gradient in each sub-regions are quantized into an eight-orientation-bin histogram weighted by the gradient variation to represent each sub-region in Fig. \ref{flowchart_descriptor}(c). To handle the bright-dark order disturbed changes, the convergence based contrast flipping estimator in \ref{flowchart_descriptor}(d) is generated. According to the sign of the estimator, either the original order version  in Fig. \ref{flowchart_descriptor}(e) or the inverted order  in Fig. \ref{flowchart_descriptor}(f) is chosen to encode the descriptor. Lastly, the histograms in each block are concatenated into one vector. Rooting algorithm is applied to further enhance the descriptor. The rooted histogram as shown in Fig. \ref{flowchart_descriptor}(g) after $L_1$ normalization forms the final DCI descriptor.

\subsection{Laplace Gradient}
	
\begin{figure}[!t]
	\centering
	\subfigure[Mean of HoG]{ \includegraphics[width=0.47\textwidth]{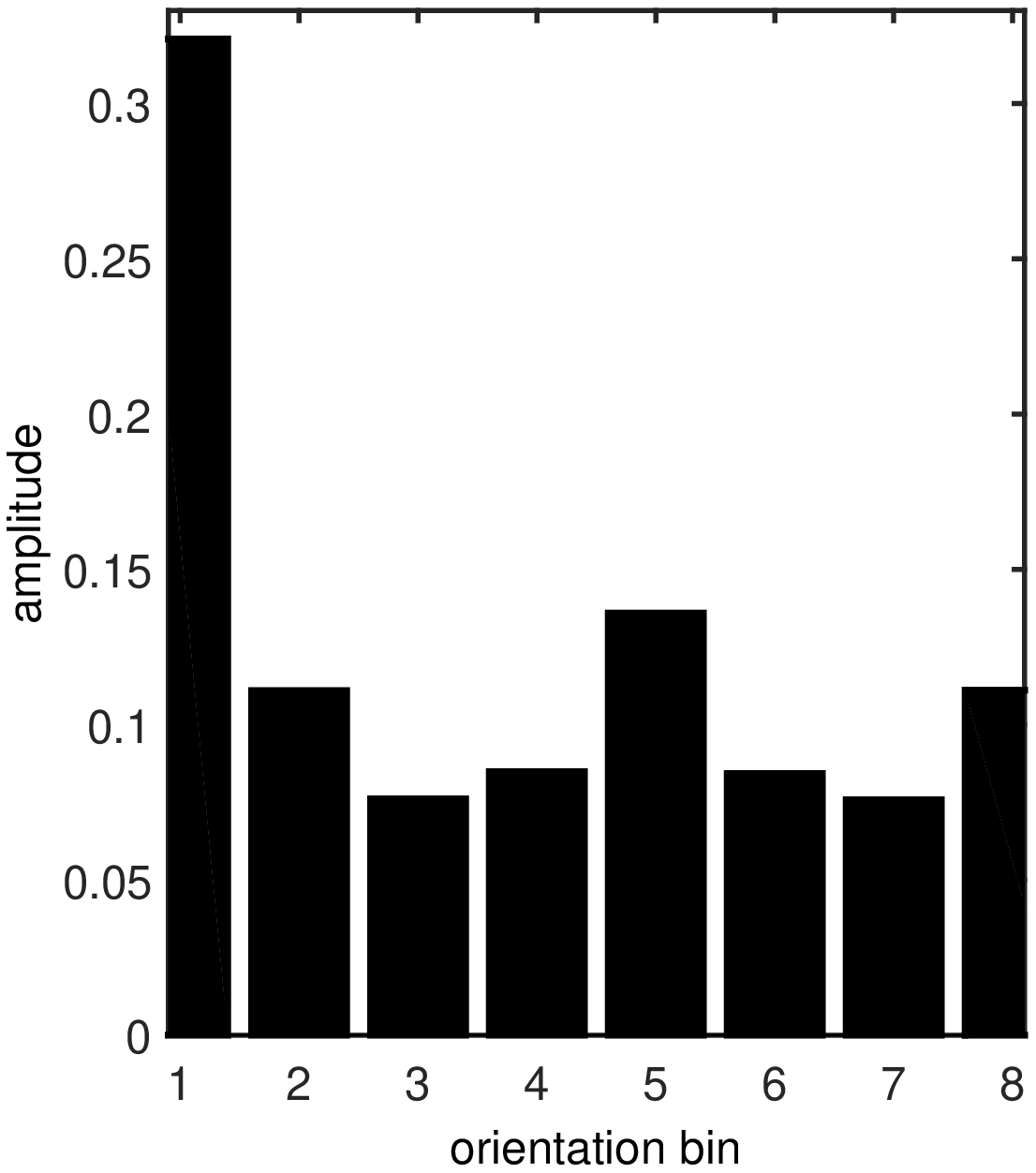} \label{HOG_mean} }
	\subfigure[Mean of HoLG]{ \includegraphics[width=0.47\textwidth]{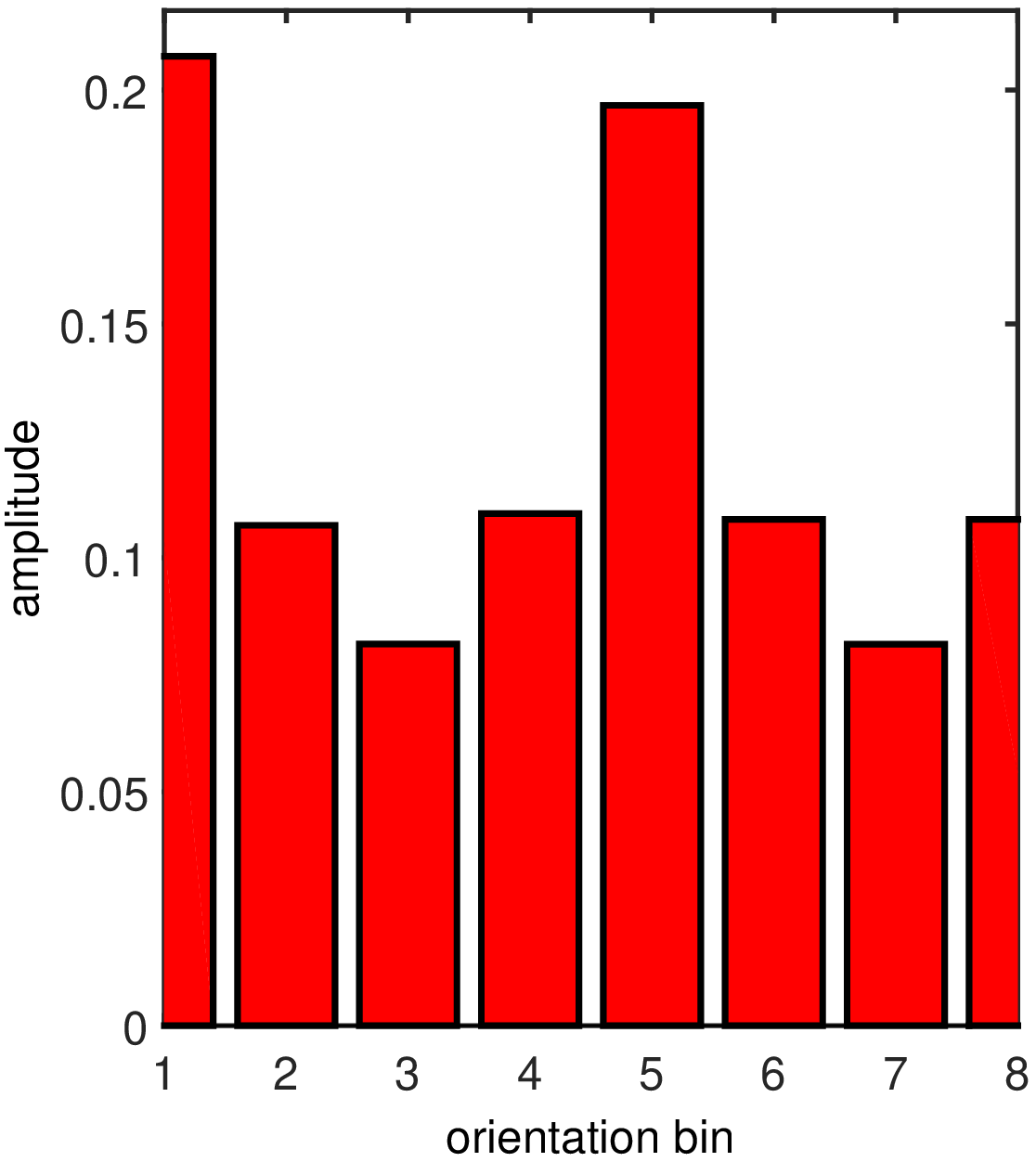} \label{HOLG_mean} }
	\caption{Mean of the eight-bin (a)  HoG and (b) HoLG over 1,500,000 interest points.}
	\label{HoG_HoGL_mean}
\end{figure}
	
The rationale to use image gradient is that it can capture both the spacial and frequency information \cite{Lowe04}. Inspired by the model of biological neuron network in which each neuron responses to a gradient with certain orientation and spatial frequency \cite{edelman1997complex}, image gradient is utilized in local descriptor. Different from the biological vision system that uses numerous neurons to process the gradient information, a local descriptor uses one statistical histogram of image gradient to describe image patchs. For changes caused by rotation, scale and viewpoints, the local patches are usually anchored by the locally detected image structures with certain image patterns (e.g., the blob detectors). Problems arise in these descriptors: 1) the alignment of image patches with the dominant orientation may reduce the discriminative ability of histogram of gradient based descriptor, and 2) the detected regions, either blobs or corners, whose shapes are similar, may reduce its discriminative ability as well. 
	
One of the solutions for addressing the image rotation variation is to use dominate orientation for aligning the image patches. Nevertheless, this alignment is criticized for the information lost, as some bins with extremely large value override the rest bins. In order to analyze the dominate orientation in a more sufficient manner, the average of the eight-bin Histogram of Gradient (HoG) over 1,500,000 local regions aligned by the dominant orientation is used. As shown in Fig. \ref{HoG_HoGL_mean} (a), the first-bin in the averaged histogram is much larger than other bins. It is about 3 times of the second bin and about 4 times of the third bin. In order to suppress the dominant bins with extreme values, hard truncation is adopted in the SIFT algorithm \cite{Lowe04}. However, a side effect of the suppression is that it will introduce nonlinear distortion, undermining the performance of the descriptor.

Another limitation of the gradient based descriptor is that the gradient field is characterized by the same pattern as the described image patch, whichever is a blob or corner. Consequently, using the gradient histogram to represent the distributions of the gradient would not be able to distinguish different images. This weakness becomes even evident in conditions when the strong illumination changes lead to great influence of both the gradient orientation and amplitude. 
	
Rather than using the first order derivatives to capture the orientation information in a local patch, the Laplace gradient uses the third order derivatives to describe the local patches. With illumination changes, the high order surface information such as the curvature is more resistant compared to the lower order information. Considering that isotropic operators are helpful in capturing the more invariant information, the Laplace gradient defined in the following is employed to describe the local patch. Let the image gradient $\mathbf{g}(\mathbf{x})$ at location $\mathbf{x}$ be 
\begin{align}
\mathbf{g}(\mathbf{x}) &= \nabla I(\mathbf{x}) \\
 &= \frac{\partial I(\mathbf{x})}{\partial x} \overrightarrow{i}+\frac{\partial I(\mathbf{x})}{\partial y} \overrightarrow{j}
\end{align}
where $\mathbf{x}=\{x,y\}$ is the coordinate, $I(\mathbf{x})$ is the pixel intensity and  $\overrightarrow{i}$ and $\overrightarrow{j}$ are the directions along $x$ and $y$ axises. 
The Laplace gradient is defined as 
\begin{align}
\label{laplace_g}
\mathbf{d}(\mathbf{x}) &= \nabla^2 \mathbf{g}(\mathbf{x})\\
& = \Big(\frac{\partial^3 I(\mathbf{x})}{\partial x^3}+\frac{\partial^3 I(\mathbf{x})}{\partial x\partial y^2}\Big) \overrightarrow{i}+\Big(\frac{\partial^3 I(\mathbf{x})}{\partial y\partial x^2} +\frac{\partial^3 I(\mathbf{x})}{\partial y^3}\Big)\overrightarrow{j}.
\end{align}
The Laplacian operator can boost the discriminative information of local features while enhancing the structures of the local patch that are characterized by the high order derivative. 
	
The implementation of the Laplace gradient is straightforward. Let choose the common used Laplacian operator
\begin{equation}
\label{laplacian_operator}
\nabla^2 = \left( \begin{array}{ccc}
0 & -1 & 0 \\
-1 & 4 & -1\\
0  & -1 & 0 \end{array} \right).         
\end{equation}
Substituting \ref{laplacian_operator} into \ref{laplace_g} yields
\begin{align}
\mathbf{d}(\mathbf{x})& = \mathbf{g}(\mathbf{x})-\frac{1}{4}(\mathbf{g}(\mathbf{x}_1)+\mathbf{g}(\mathbf{x}_2)+\mathbf{g}(\mathbf{x}_3)+\mathbf{g}(\mathbf{x}_4)),
\end{align}
where  $\mathbf{x}_1$, $\mathbf{x}_2$, $\mathbf{x}_3$ and $\mathbf{x}_4$ are the adjacent horizontal and vertical neighborhoods cells of $\mathbf{x}$.

\begin{figure}[t]
	\centering
	\includegraphics[width=0.9\textwidth]{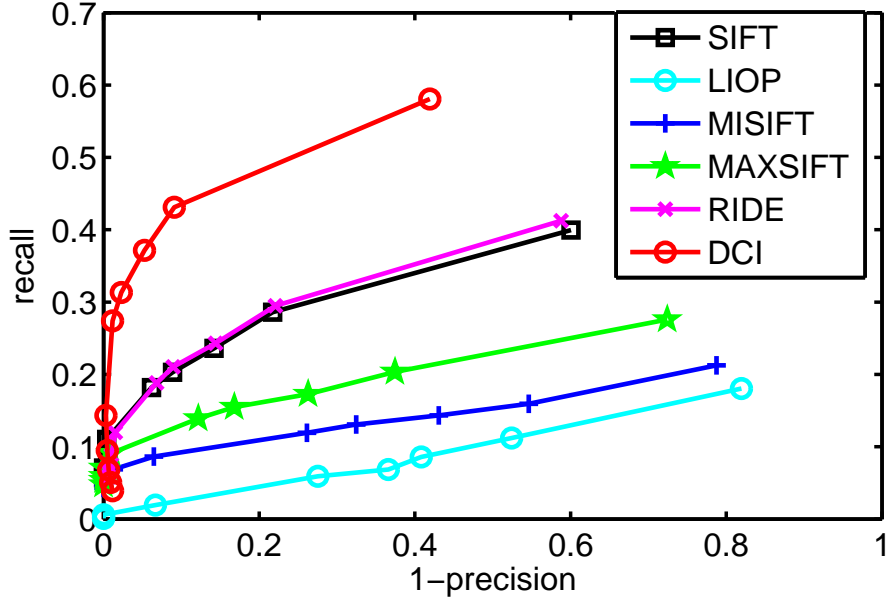}
	\caption{Performance comparison of the SIFT, LIOP, MISIFT, MAXSIFT, RIDE and DCI descriptors under the large illumination changes.}
	\label{precision_recall_on_wallpaperimages}
\end{figure}

From the Laplace gradient, eight-bin Histogram of Laplace Gradient (HoLG) is formed to represent the corresponding blocks as shown in Fig. \ref{flowchart_descriptor}(c). The average of the HoLG over 1,500,000 local regions is shown in Fig. \ref{HoG_HoGL_mean} (b). It is suggested that the HoLG alleviates the overriding-bin problem compared to that of the HoG. In order to further test the performance of the HoLG, a image pair as shown in Fig. \ref{products_sample_images} is selected. Take the first and the third images in the third row. These two images are in the same sketch form yet in different painting color. Their intensity values of the corresponding gray images differ from each other a lot, leading to a dramatically decrease in the descriptor matching rate. The recall-precision  \cite{Mikolajczyk05} which is calculated by the ratio of the number of correctly matched point pairs over the incorrectly matched point pairs is employed here to evaluate the performance of the descriptors. The matching strategy considers the nearest neighbor distance ratio and declares a match if the distance ratio between the first and the second nearest neighbors is below a threshold. The number of correct matches and ground truth correspondence is determined by the overlap error. A match is correct if the overlap error$<0.5$. The results are presented with \emph{recall} versus 1-\emph{precision} curves
\begin{equation}
recall=\frac{\#correct~matches}{\#correspondences},
\end{equation}  
\begin{equation}
1-precision=\frac{\#false~matches}{\#all~matches},
\end{equation} 
where $\#correspndeces$ is the ground truth number of matches. As shown in Fig. \ref{precision_recall_on_wallpaperimages}, the DCI descriptor implemented on the basis of the Laplace gradient achieves a better recall-precision under the large illumination changes.  From this we can find that the HoLG based DCI descriptor has a drastically better performance than the HoG based descriptors such as SIFT, MAXSIFT, LIDE, MISIFT and the intensity order based descriptor LIOP.
	
\subsection{Divergence based Contrast Flipping}
As for the bright-dark flipping issue, a simple way to consider is reversing the contrast of the whole image to the opposite intensity. However, as suggested  in \cite{xie2015ride}, converting the whole images to the inverted contrast is only optimal to the global image contrast inversion. If only parts of the image's contrast are inverted, the flipping approach is not applicable. The MI-SIFT uses an average of both the SIFT before and after flipping. However, the discriminative ability is comprised by this way. Alternatively, a criterion is set to determine whether a certain region should be flipped or not. For instance, the MAX-SIFT takes the maximun alphabetical order of the two transformed version as the criterion. However, the results are subject to noises.
	
In view of previous limitations, we propose to use the divergence of the local region to determine whether the local region is bright or dark. A bright blob is a region that the gradient converges to the center whereas a dark blob is the region that the gradient diverges to the center. The divergence of gradient is defined as 
\begin{align}
\textrm{div} \mathbf{g}(\mathbf{x})& = \nabla \mathbf{g}(\mathbf{x})\\
& =\nabla^2 I(\mathbf{x}).
\end{align}
The equation indicates that the divergence of the gradient is the Laplacian of the image. Thus, it is plausible that the blob can be detected by the response of the Laplacian filter. The response of the Laplacian filter produces the largest signal to noise ratio at the center of the blob region because its shape matches with that of the blob. Therefore, it is conceivable that the divergence is more reliable in detecting whether this region is bright or dark. The surface integration of the divergence is 
\begin{equation}
\varPhi(\mathbf{x}) = \iint_s \textrm{div} \mathbf{g}(\mathbf{x}) ds. 
\end{equation}
The implementation of the surface integration is clear. It can be easily computed from the gradient. Based on the Stokes' theorem \cite{springer1957introduction}, the integration of the divergence gradient over a surface equals to the integration of the gradient over the boundary of the local region, that is
\begin{align}
\varPhi(\mathbf{x}) & = \iint_s \textrm{div} \mathbf{g}(\mathbf{x}) ds \\
& = \oint_l \mathbf{g}(\mathbf{x})\cdot \mathbf{n}dl,
\end{align}
where $\mathbf{n}$ is the outward-pointing unit normal vector on the boundary.

\subsection{The DCI Descriptor}
	
With the Laplace gradient, an additional step to tackle the illumination changes is adopting the rooting algorithm \cite{Arandjelovic12}. Although $L_1$ normalization is carried out for the concatenation of the histograms from the Laplace gradient, the normalization cannot effectively respond to the problems caused by the nonlinear illumination changes. Inspired by the work in \cite{Arandjelovic12}, the rooting algorithm, instead of the hard truncation threshold \cite{Lowe04}, is incorporated by simply taking the square root of descriptor value in each dimension. 
	
As a summary, the DCI descriptor is derived as follows. Let the concatenation of the Laplace gradient in all the $4\times4$ sub-regions be $ H =\{ h_1,..., h_{128}\}$. The $L_1$ normalized $ H$ is 
\begin{equation}
H_n =\{ h_{1n},..., h_{128n}\}
\end{equation} where 
\begin{equation}
h_{in} = \frac{ h_{i}}{\sum{H}}.
\end{equation}
 The DCI descriptor $H_{nr}$ is the square root of the elements in $H$, defined as 
\begin{equation}
H_{nr} =\{\sqrt{h_{1n}},...,\sqrt{h_{128n}}\}.
\end{equation}	
The performance of the DCI descriptor will be evaluated in the following section.

\section{Experiments}
In this section, we evaluate our proposed DCI descriptor with comparison to the state-of-the-art SIFT \cite{Lowe04}, LIOP \cite{Wang11},  MISIFT \cite{ma2010mi}, MAXSIFT \cite{xie2014max} and RIDE \cite{xie2015ride} descriptors in the applications of logo visual search, wallpaper visual search, and face recognition.

\subsection{Logo Visual Search}
\begin{table*}[!t]
	\caption{Retrieval Performance on the BelgaLogos Database.}
	\label{tab_logo}
	\begin{center}
		\begin{tabular}{@{\hspace{\tabcolsep}%
						\extracolsep{\fill}}ccccccccc} \hline
			Logos		    &SIFT  		  & LIOP & MI-SIFT & MAX-SIFT & RIDE  & DCI \\\hline
			\includegraphics[height=0.018\linewidth]{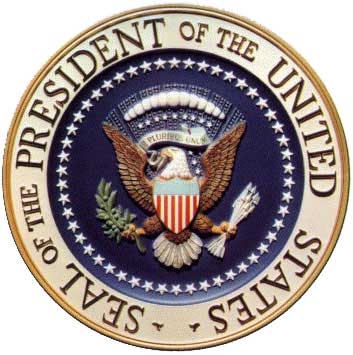}     & 100\%       &100\%       &87.3\%     &92.9\%  & 100\%        		& \textbf{100}\%    \\\hline 
			\includegraphics[height=0.018\linewidth]{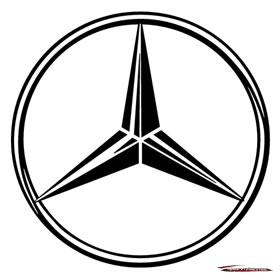}   	        & 22.0\%      &4.3\%       &27.9\%     &25.8\%  & 39.8\%       		& \textbf{40.4}\%    \\\hline
			\includegraphics[height=0.018\linewidth]{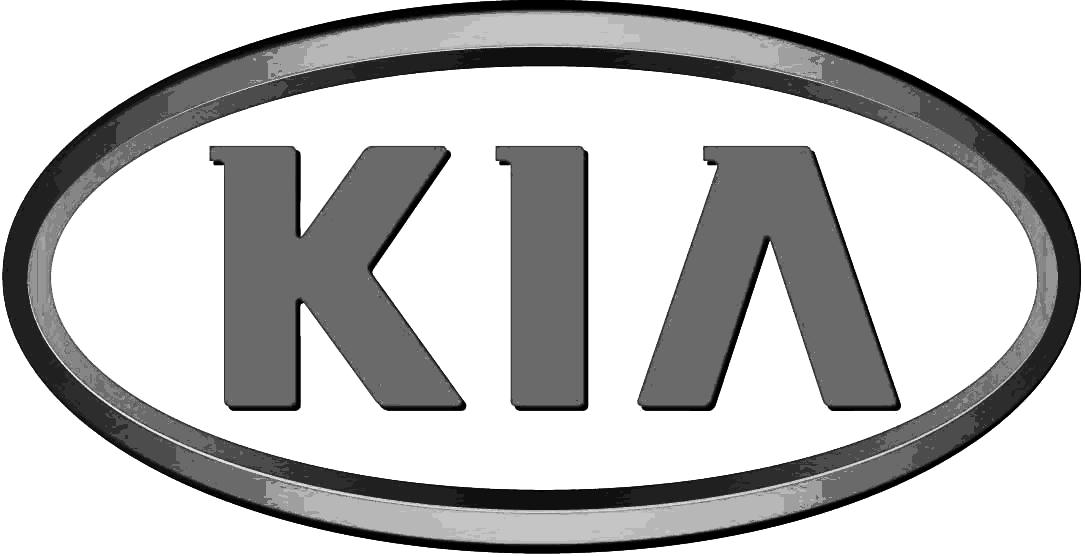}   	            & 61.9\%      &37.4\%      &60.7\%     &66.6\%  &{68.5}\%    & \textbf{68.8}\%      		\\\hline
			\includegraphics[height=0.018\linewidth]{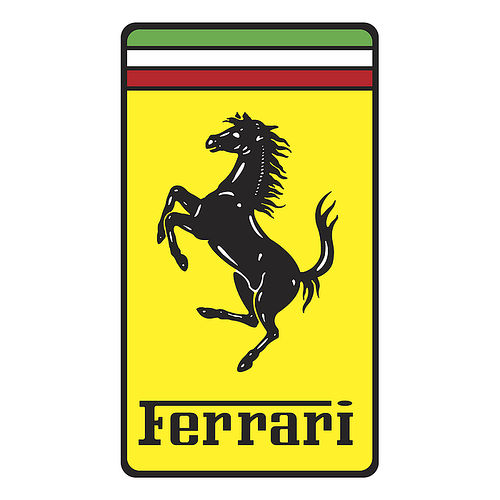}               & {24.6}\%    & 5.7\%      & 8.1\%     & 10.8\% & 24.3\%       		& \textbf{31.6}\%  	\\\hline
			\includegraphics[height=0.018\linewidth]{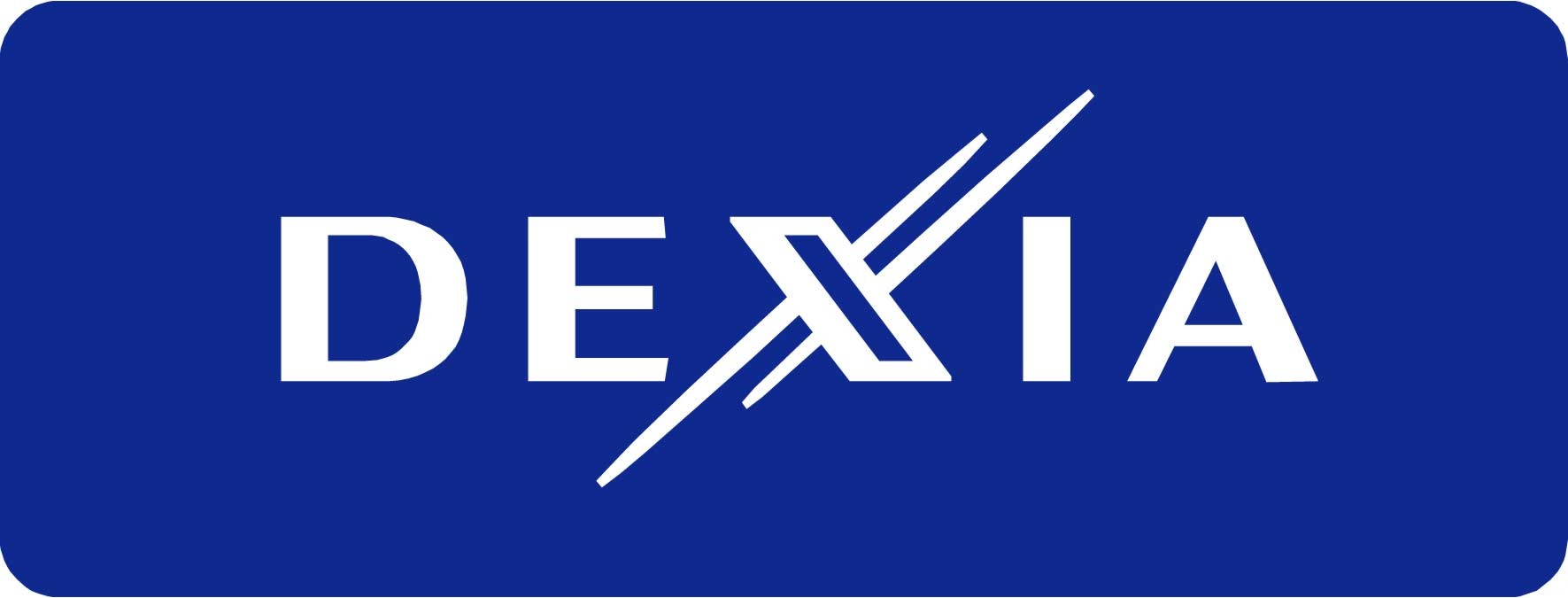}                 & {68.8}\%    & 49.3\%     & 82.5\%    & 85.6\% & 87.0\%        	& \textbf{87.7}\%  	\\\hline
			\includegraphics[height=0.018\linewidth]{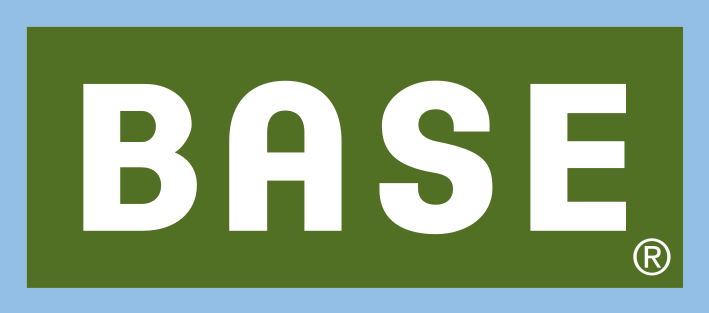}                  & \textbf{67.8}\%    & 51.1\%     & 61.0\%    & 66.3\% & 67.5\%        	& {66.5}\%  \\\hline
			mAP                                                                 & {57.5}\%    &41.3\%      & 54.6\%    &57.8\%   & 64.5\%              & \textbf{65.8}\%  	\\\hline
		\end{tabular}
	\end{center}
\end{table*}
	
\begin{table*}[!t]
	\caption{Average Precision for the 'Kia' Logo with different rotation on the BelgaLogos Database.}
	\label{tab_logo_rotation}
	\begin{center}
		\begin{tabular}{@{\hspace{\tabcolsep}%
						\extracolsep{\fill}}cccccccc} \hline
			Rotation Degree		    & SIFT 		  & LIOP & MI-SIFT & MAX-SIFT & RIDE  & DCI \\\hline
				90                      & 61.4\%      &35.0\%  &63.4\%    &65.8\%  & 61.8\%        & \textbf{68.2}\%  \\\hline 
				180    	                & 62.3\%      &36.4\%  &62.0\%     &65.0\%  & 22.7\%        & \textbf{67.0}\%  \\\hline 			
			\end{tabular}
\end{center}
\end{table*} 	

\begin{figure}[t]
	\centering
	\includegraphics[width=0.77\textwidth]{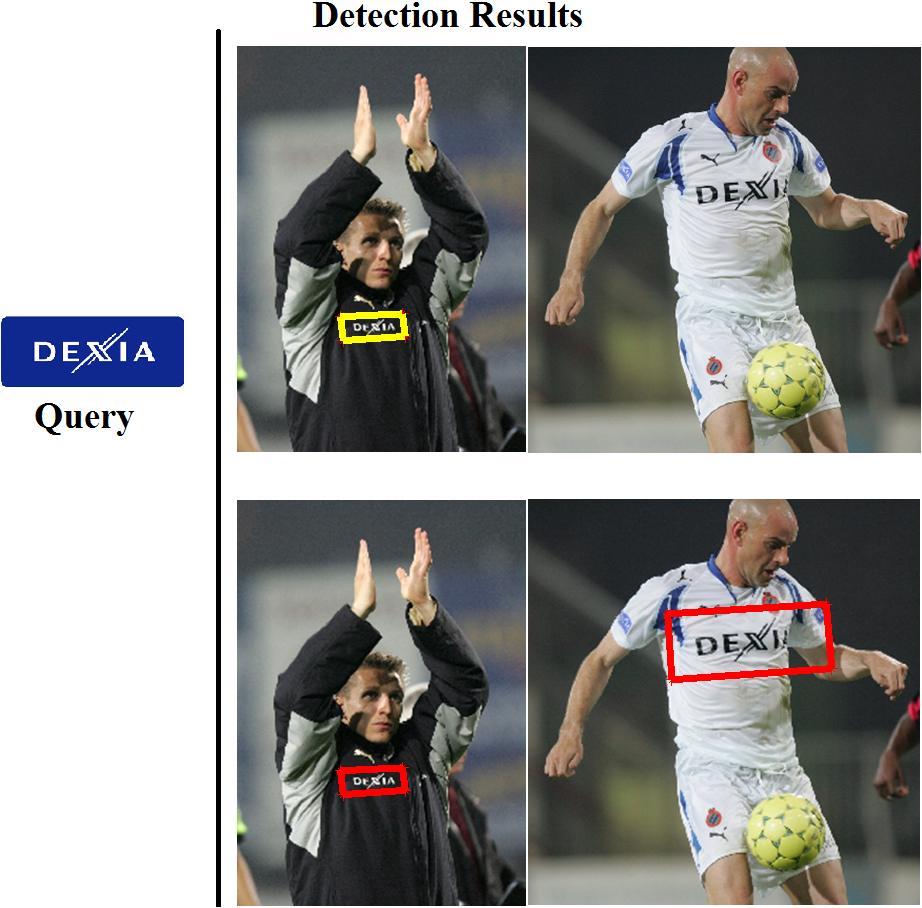}
	\caption{Logo detection results based on the SIFT and DCI descriptors. The first row shows the results obtained using the SIFT descriptor marked with yellow bounding box, and the second row illustrates the results obtained using the proposed DCI descriptor marked with red bounding box. The proposed DCI can effectively detect both two logos in these two images but SIFT cannot.}
	\label{logo_results}
\end{figure}
	
First, we test the proposed DCI descriptor on the challenging BelgaLogos database \cite{Joly09}, which contains 10,000 real-world images from various events. Samples are shown in the forth row of Fig.\ref{products_sample_images}. The image size is kept the same as that provided in \cite{Joly09}. Hessian-Laplacian detector \cite{mikolajczyk2004scale} with the default setting from vl-feat\footnote{The vl-feat is downloaded from http://www.vlfeat.org/} is used to extract the interest points from each image. The brute-force search algorithm is used for all the descriptors evaluated here: the DCI, MISIFT, MAXSIFT, RIDE, LIOP, and SIFT descriptors. The evaluation is tested on the widely evaluated six logos \cite{jiang2012randomized}, the `US-President', `Mercedes', `Kia', `Ferrari', `Dexia' and `Base'. 
	
The Average Precision (AP) of each logo and the mean Average Precision (mAP) for all six logos are given in Table \ref{tab_logo}. It shows that the DCI descriptor outperforms other descriptors most of the cases. It increases the mAP by more than 8\% compared to the SIFT descriptor. In order to explain the performance improvement of the DCI, visual inspections on the logo detection are given in Fig. \ref{logo_results}. It shows that the SIFT descriptor can detect the logo in the top left image but the top right image where the bright and dark parts are the inverted of query. As expected, our proposed DCI descriptor can effectively detect the logo from both images under such severe contrast changes. 
	
Experiments are also carried to test the descriptors under image rotation condition. The 'Kia' logo with the rotation of 90 degree and 180 degree are given in Table \ref{tab_logo_rotation}. It shows that DCI outperforms others under the image rotation while RIDE is sensitive to the rotation.

\begin{table*}[!t]
	\caption{Retrieval Performance on the Wallpaper Database.}
	\label{tab2}
	\begin{center}
	\begin{tabular}{@{\hspace{\tabcolsep}%
							\extracolsep{\fill}}cccccccc} \hline
		& SIFT             & LIOP               & MI-SIFT          & MAX-SIFT  & RIDE    & DCI  \\\hline 
		mAP (\%)   	                 & 64.0           &  49.5             & 61.5       &  67.6   & 64.5        & \textbf{74.0}  \\\hline 
	\end{tabular}
	\end{center}
\end{table*}

\begin{figure}[!t]
		\centering
		\subfigure[]{ \includegraphics[width=0.6\textwidth]{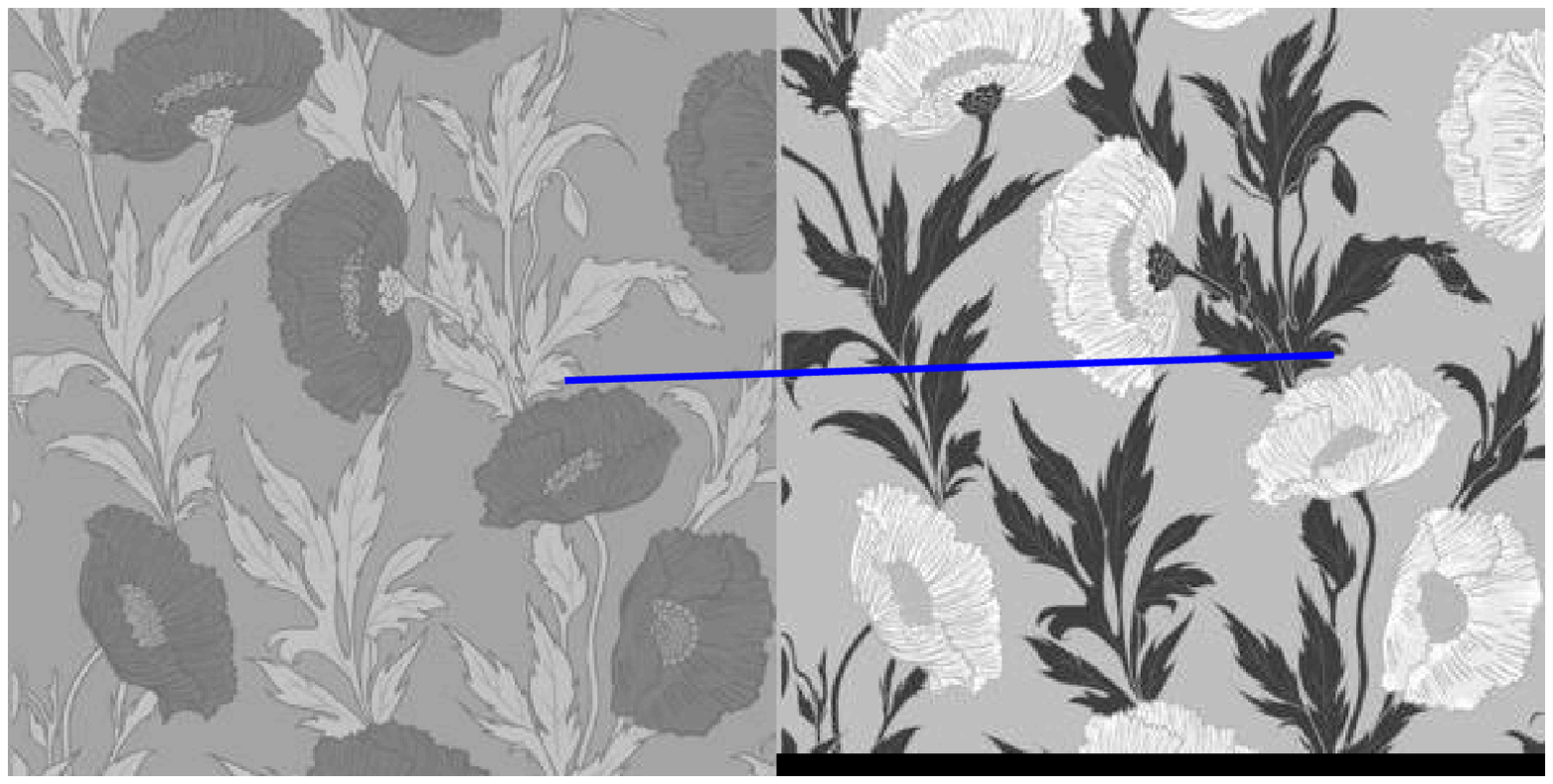} \label{SIFT_default} }
		\subfigure[]{ \includegraphics[width=0.6\textwidth]{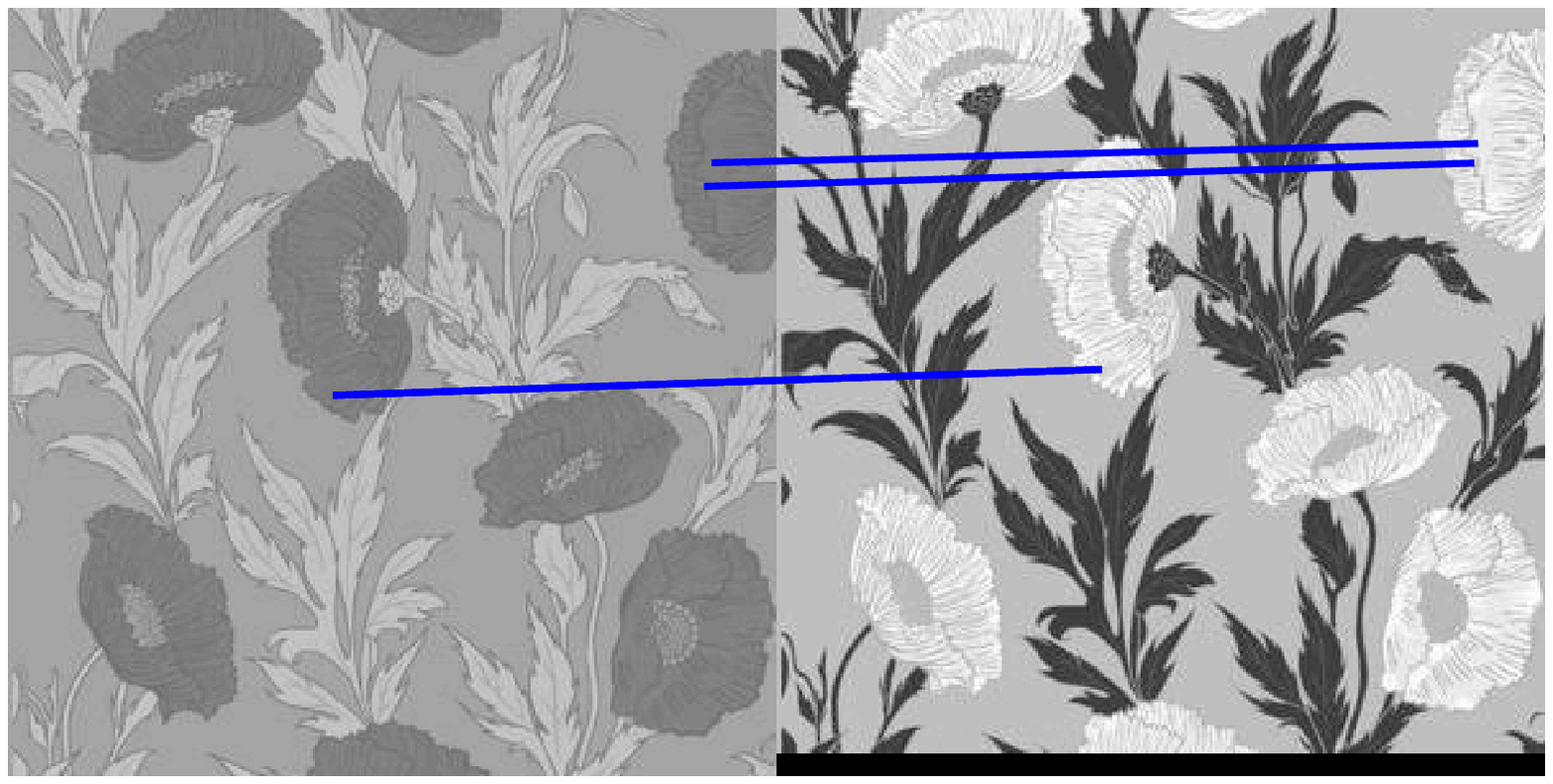} \label{LIOP_example} }
		\subfigure[]{ \includegraphics[width=0.6\textwidth]{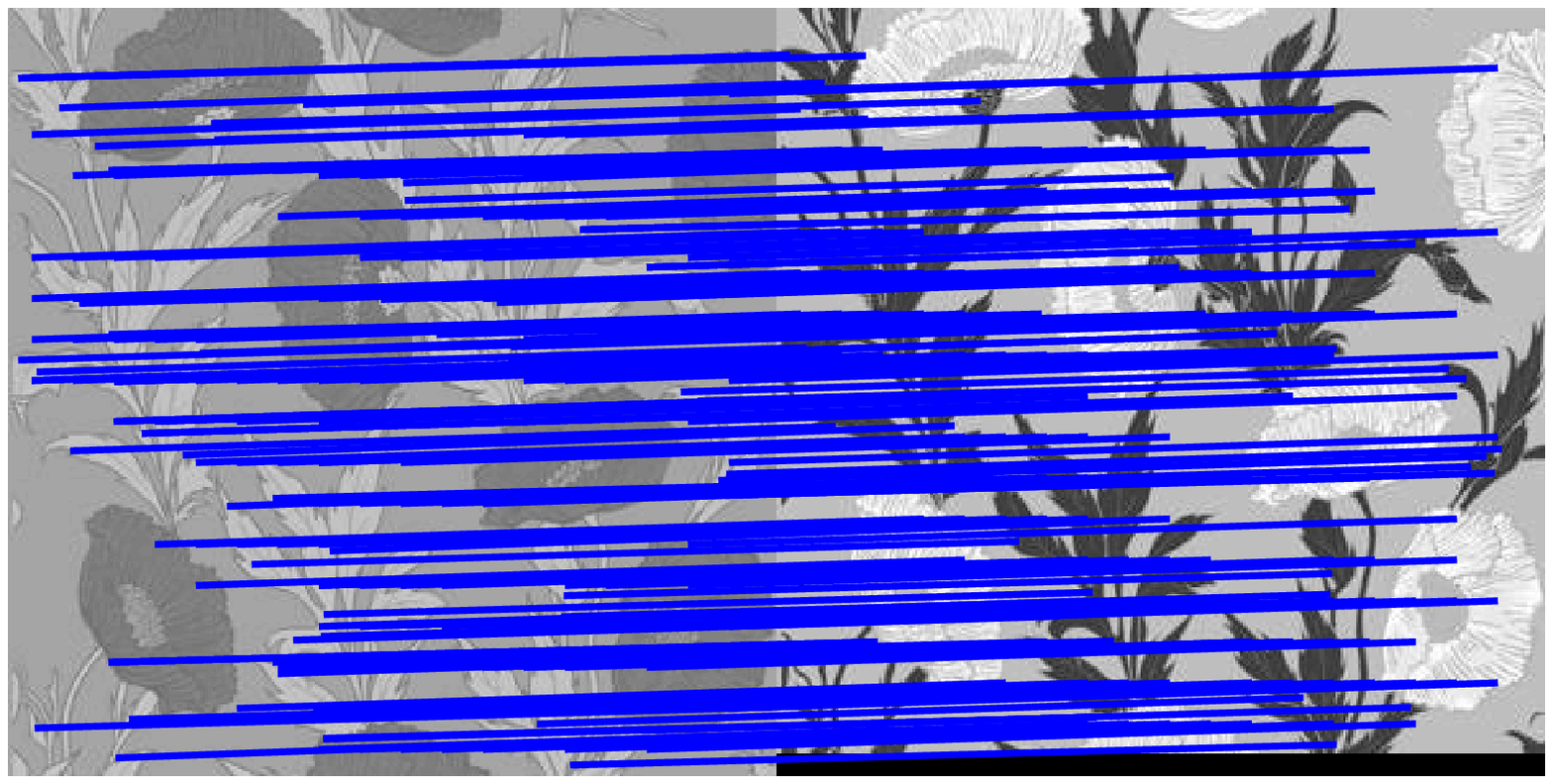} \label{SIFT_proposed} }
		\caption{Correct matched pairs obtained by using (a) SIFT, (b) LIOP, and (c) DCI descriptors. The blue lines represent the correctly matched pairs between the query and reference images.}
		\label{visual_exp}
	\end{figure}

	\subsection{Wallpaper Visual Search} 
	\label{a}
	The reason we choose wallpaper as a test set is that it includes a considerable number of images with various illumination variations from the same design. Samples of a wallpaper design (which is also named as category in the following) are shown in the third row of Fig. \ref{products_sample_images}. In total, the reference image set contains 522 images from 77 categories which are provided by a wallpaper design company. The test image set contains 1,014 images.

	The wallpaper search algorithm in \cite{Yap15} is implemented in this. Specifically, the hierarchical k-means clustering is used to train the Scalable Vocabulary Tree (SVT) \cite{NisterCVPR06}. A branch factor of 10 and depth of 6 are set for the SVT. A total of 1,000,000 words are trained for the quantization. While keeping the original aspect ratio, the images are normalized to the standard size with the longest dimension as 640 pixels. The interest point detector is a combination of the dense and the adaptive sparse SIFT which is the same as that in \cite{Yap15}. Both sparse and dense local interest points are used to anchor the local features. The mAP is calculated to indicate the retrieval performance.

	The image retrieval  results in Table \ref{tab2} show that the proposed DCI improves the retrieval performance by 6\%  compared with the other descriptors. It demonstrates that the DCI is more robust to the severe contrast changes which widely exist in the wallpaper datasets.
	
As shown in Fig. \ref{visual_exp}, eyeballing two images with large illumination changes can find that both the SIFT and LIOP descriptors identify only limited number of correctly matched pairs in the event of severe illumination changes. In contrast, the proposed DCI descriptor can identify much more correctly matched pairs, suggesting its superior performance to other evaluated descriptors. 

	\begin{figure}[t]
		\centering
		\includegraphics[width=0.9\textwidth]{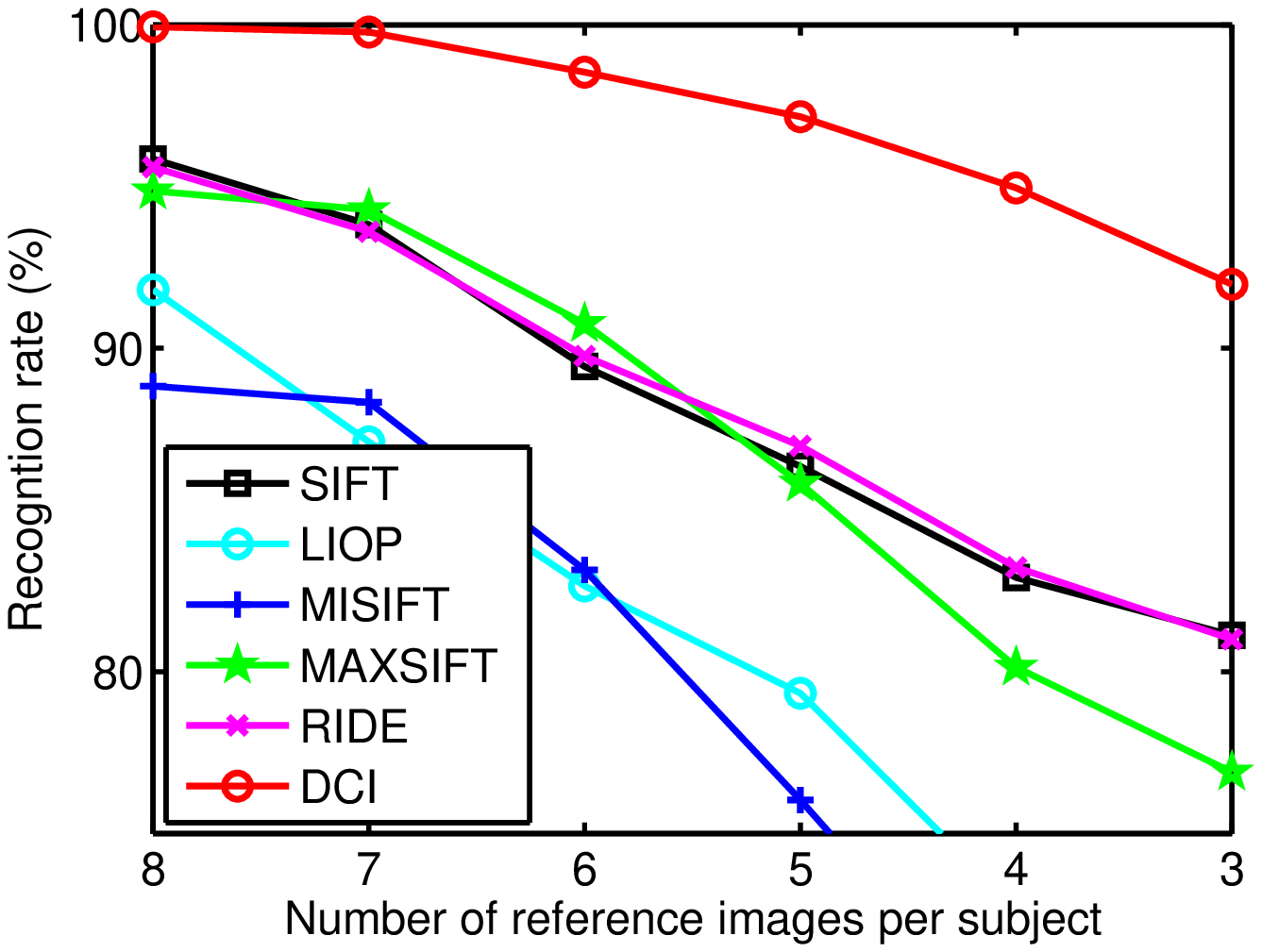}
		\caption{Face recogntion rate versus number of reference images on the Multi-PIE database.}
		\label{Face_recognition}
	\end{figure} 
		
	\subsection{Face Recognition}
	
	Over years, face recognition is always an active research topic \cite{Jiang08,Miao08,miao2009human}. Interest points are also applied to face recognition \cite{Geng13}. Many face databases are publicly available, providing a convenient and rich test set for evaluation.  In this experiment, we test the proposed DCI descriptor on one of the widely accepted face databases: the CMU Multi-PIE database \cite{Gross10}. 
	
	The CMU Multi-PIE database contains face images captured in 4 sessions with variations in illumination, expression and pose. For the purpose of this experiment, the face image sets with illumination variation is selected. The first 105 subjects that appear in all 4 sessions are used. Images are cropped  and down-sampled to the size of $100\times82$ pixels.

	As illustrated in \cite{Gross10}, the illumination variation dataset contains 18 flash-only images and 2 non-flash images per person. Samples are shown in the second row of Fig. \ref{products_sample_images}. In total, 20  neutral-expression images with different illumination are used for the evaluation, which produce $20\times4\times105 = 8400$ images for the evaluation. For each subject, the first $t$ images are selected as the reference and the rest $20-t$ images are used as the query. The $\ell_0$-LoG detector \cite{Miao15TIP} which is an illumination invariant interest point detector is used to extract the keypoints from the face images. The algorithm of image recognition through interest point matching is adapted from \cite{Lowe04}. Experimental results are shown in Fig. \ref{Face_recognition}. It shows that the proposed DCI descriptor outperforms other descriptors over all cases. When the reference number of images is 3, the DCI descriptor achieves a 15\% incremental compared to other descriptors.  Again, it suggests that the proposed DCI descriptor outperforms other descriptors on the face recognition experiment.

\section{Acknowledgements}	
This work was supported in part by the Media Development Authority, Singapore, under Grant MDA/IDM/2012/8/8-81 Vol01, in part by the Rapid-Rich Object Search Laboratory, Nanyang Technological University, Singapore, and in part by the National Research Foundation, Prime Ministers Office, Singapore, under its Interactive and Digital Media (IDM) Futures Funding Initiative and administered by the IDM Programme Office.

\section{Conclusions}	
\label{Sec:conclusion}

In this work, we propose a local feature descriptor named DCI that is discriminative and contrast invertible in illumination changes and contrast inversion. The Laplace gradient is computed to describe each pixel. A divergence-based contrast flipping estimator is created for images with the bright/dark disturbed variations. The square root of the HoLG after $L_1$ normalization is used to further mitigate the problems caused by illumination changes. Experiments on the BelgaLogos, Wallpaper, Multi-PIE databases exhibit the superior performance of the DCI descriptor in object search applications over the state-of-the-art descriptors. It suggests that the DCI descriptor is robust to the illumination changes and contrast inversion.





\bibliographystyle{elsarticle-num}







\end{document}